\documentclass[letterpaper, 10 pt, conference]{ieeeconf}

\IEEEoverridecommandlockouts 

\usepackage{multirow}
\usepackage{cite}
\usepackage{graphicx}
\usepackage{amsmath}
\usepackage{algorithmicx}
\usepackage{algpseudocode}
\usepackage{array}
\usepackage[caption=false,font=footnotesize]{subfig}

\begin{document}
\title{\LARGE \bf Optimal Alarms for Vehicular Collision Detection}

\author{Michael Motro, Joydeep Ghosh, and Chandra Bhat%
\thanks{The authors are with the University of Texas at Austin, 78712 TX, USA.}}



%


\maketitle

\begin{abstract}
An important application of intelligent vehicles is advance detection of dangerous events such as collisions. This problem is framed as a problem of optimal alarm choice given predictive models for vehicle location and motion. Techniques for real-time collision detection are surveyed and grouped into three classes: random Monte Carlo sampling, faster deterministic approximations, and machine learning models trained by simulation. Theoretical guarantees on the performance of these collision detection techniques are provided where possible, and empirical analysis is provided for two example scenarios. Results validate Monte Carlo sampling as a robust solution despite its simplicity. 
\end{abstract}


%
\IEEEpeerreviewmaketitle

\section{Introduction}
Recent advances in in-vehicle awareness have end uses such as messages or warnings to drivers, automated braking or control, or fully driverless vehicles. There are similarly many sensors and communication devices that can provide awareness, and many models of traffic motion or human action that add predictive power. As there are many possible approaches, a single unified framework for intelligent vehicle design seems unlikely in the near future.
However, there are certain tasks that are important for a variety of intelligent vehicle applications and (relatively) independent of the individual sensors or models used. One such task is vehicular collision detection: given the current position and state of two or more vehicles and a predictive model for their future motion, determine whether there is a significant chance of collision between vehicles in the near future. This task may sound trivial and is indeed simpler than the problems of scene reconstruction, predictive modeling or path planning. This simplicity allows vehicular collision detection to be framed as a self-contained task, with solutions that compromise between speed and robustness. \par
Collision detection closely matches the theoretical problem of optimal alarm design \cite{demare, bayesevent}. Optimal alarms were initially studied in the context of detecting bankruptcies or machine part failures \cite{conditioncasestudy} -- critical events that should be detected in advance with high probability, much like collisions. The basic mathematical framework of optimal alarm theory is as follows: let $t=0$ be the current time, and $X_0$ represent all information known about the studied system at this time. The random variable $X_t$ represents the system’s state at time $t$, and there is a predictive model $P(X_t | X_0)$ for all $t$. The probabilistic event $C$ is defined as a known set of events in $X_t$ that share some critical property, for instance the event that two vehicles have collided at any time $t$. 
An alarm $A$ is a true or false action that corresponds with the prediction of event $C$. For vehicular collision detection, this alarm may represent a warning to the driver, the activation of an emergency program, or simply a notification to another part of the autonomous driving system. As an example of the latter, some motion planning techniques will develop several candidate motions, then remove candidates that are likely to result in collisions \cite{rrt1, worst1}. If the future can be known perfectly, then the optimal alarm is $A = C$. In reality there will be some probability of false negative alarms, where a collision will occur but no alarm was sent, or false positive alarms, where an alarm is sent but there is no collision. The goal of an alarm algorithm is to efficiently calculate $A$ so as to minimize false negatives and positives.\par
Unfortunately, most of the previously developed optimal alarm techniques concern events that are mathematically simple to characterize, such as bankruptcy (money $< 0$), and require predictive models with convenient properties. Modern models for object detection, localization, and vehicular motion prediction are varied and complex. Collision detection requires an estimate or probability distribution of each vehicle's current position, as well as a transition function that gives the probability of a vehicle moving from one state to another in a fixed time. Popular vehicle tracking techniques, such as Kalman filters, particle filters, and autoregressive models, provide this information. \cite{survey1} surveys vehicular motion prediction models and discusses their use in collision detection. Additionally, the mathematical formulation for collision between two vehicles, while certainly tractable, generally requires multiple non-linear operations and thus is more complex than the critical event of traditional alarm systems. As such, we focus on techniques developed specifically for vehicular collision detection. \par 
Section II develops the mathematical framework for vehicular collision detection and discusses a useful performance metric. Section III describes existing techniques for collision detection, providing guarantees on average or worst-case performance where possible. Section IV compares these techniques with two simulated example scenarios.

\section{Formulating Vehicular Collision Detection}
\subsection{Defining the Collision Event}
Vehicles are defined as shapes on a 2-D grid whose coordinates, orientation, and even dimensions may be partially random. The state $X_t$ is a random variable that contains all knowledge of the environment at each time $t$. For instance, say $X^v_t$ represents vehicle $v$'s current position and dynamics at time $t$, as well as other measurable or hidden variables including human factors. Then $X_t$ is the product of all these states $\left\langle X^1_t,X^2_t,\cdots\right\rangle$. A motion model $P(X_{t+1} | X_t)$ is assumed to be known. 
If the space occupied by two vehicles overlaps at a time $t$, then these vehicles are considered in collision and the event $C_t$ is true. There are well-known methods to determine overlap between two simple shapes such as rectangles or ellipses, and more complex vehicle shapes can be described as a combination of rectangles and ellipses or approximated in other ways \cite{collisiondetection}.
\par 
An alarm predicts a single event that is a function of all $C_t$. We focus on the event $C = \bigcup_{t=0}^{t_f} C_t$ for some cutoff time $t_f$, in other words the event that a collision will have occurred by time $t_f$. 
Alarms that denote the immediacy, in addition to the probability, of a collision situation use values known as criticality measures \cite{ttx1}. The most common criticality measure is time-to-collision, typically defined as $TTC = \inf \{t : C_t = \textnormal{True}\}$. Time-to-collision cannot be directly determined from the collision event defined above, but can be constructed using multiple alarms with different cutoff times as $TTC = \inf \{t_f: \bigcup_{t=0}^{t_f} C_t = \textnormal{True}\}$ \cite{bayesthesis}.
\par
A continuous-time version of the collision event $C$ can be calculated if vehicle shapes and motion are properly restricted as in \cite{other7, mean2}. However, for the wider class of predictive models, the only choice is to check for collision at a discrete series of time points. This causes false negative errors in the case that two vehicles collide between two checked time points, but are not in collision at those time points. These errors can be mitigated by decreasing the time interval between collision checks or by adding a margin around each vehicle, which increase the computation time or the false positives respectively.

\subsection{Quantifying Alarm Performance}
The quality of an alarm is determined by the probability of two types of error. $FP(X_0) = P(A,\bar{C} | X_0)$ is the probability of a false positive, or an alarm being sent at state $X_0$ but no collision occurring in the near future. $FN(X_0) = P(C,\bar{A} | X_0)$ is the probability of a false negative at $X_0$, or a collision occurring without a corresponding alarm. Given the probability that each state $X_0$ occurs, $P(X_0)$, then the total system's false positive and negative probabilities are $FP = \int FP(X_0) P(X_0) dX_0$ and $FN = \int FN(X_0) P(X_0) dX_0$. Metrics such as the sum of $FP+FN$ or $\max(FP,FN)$ give equal importance to the two error types, whereas graphical representations like the receiver operating characteristic show a variety of tradeoffs between them. For the purposes of collision detection it is assumed that false negatives are significantly more problematic than false positives. Additionally, the distribution of initial states $P(X_0)$ is unknown and may vary greatly depending on the application. A scoring metric that is relatively independent of this distribution is desirable.
\par
\begin{table}[!t]
\caption{Cost of Alarm System}
\label{cost_table}
\centering
\begin{tabular}{cccc}
 & & \multicolumn{2}{c}{Collision $C$} \\
 & & True & False \\
\multirow{2}{*}{Alarm $A$} & True & $0$ & $R_{FP}$ \\
 & False & $R_{FN}$ & $0$
\end{tabular}
\end{table}
A simple way to measure alarm performance is to assign a cost of $R_{FN}$ to each false negative and $R_{FP}$ to each false positive event, as shown in Table \ref{cost_table}. These costs can be determined based on the ultimate usage and effect of the alarm. For instance, one might consider an unnoticed collision 10 times more problematic than a false alarm. Use $EC_A(X_0)$ to denote the expected cost of alarm $A$ for the initial state $X_0$.
\begin{equation} EC_A(X_0) = R_{FN}FN(X_0) + R_{FP}FP(X_0) \end{equation}
The same expected cost can easily be generalized across the system.
\begin{equation} EC_A = \int EC_A (X_0) P(X_0) dX_0 = R_{FN}FN + R_{FP}FP \end{equation}
This metric allows a user to quantify the relative importance of missed alarms over false alarms. Note that if the quantity $P(C|X_0)$ is known directly, the expected cost can be rewritten as:
\begin{equation*}
EC_A(X_0) = \left\{ \begin{array}{l|r} R_{FN}P(C|X_0) & A=\textnormal{False} \\ R_{FP}(1-P(C|X_0)) & A=\textnormal{True} \end{array}\right\}
\end{equation*}
The optimal alarm given perfect knowledge of the probability of collision is then
\begin{align}
A_O = \textnormal{True if } P(C|X_0) > c_{cut} \nonumber \\
c_{cut} = \frac{R_{FP}}{R_{FN}+R_{FP}} \nonumber \\
EC_{A_O}(X_0) \leq \frac{R_{FP} R_{FN}}{R_{FN}+R_{FP}}
\end{align}

\subsection{Approximate Alarms}
An actual alarm typically uses an approximation of $P(C|X_0)$. This approximate alarm $A_\epsilon$ can be generalized as using an estimate of $P(C|X_0)$ that lies in a confidence interval of size $\epsilon$ and confidence $P_\epsilon$.
\begin{align}
&A_\epsilon = \textnormal{True if } \hat{c} > c_{cut} \nonumber \\
&P( \left| \hat{c} - P(C|X_0) \right| > \epsilon) < P_\epsilon 
\end{align}
The performance of an approximate alarm can best be determined by direct comparison to the optimal alarm. Define the Expected Additional Cost of an alarm as
\begin{align}
&EAC_A(X_0) = EC_A (X_0) - EC_{A_O} (X_0) \nonumber \\
&EAC_A = EC_A - EC_{A_O}
\end{align}
The Expected Additional Cost of the approximate alarm $A_\epsilon$ can be bounded for several cases of $P(C|X_0)$.
\begin{align*}
&P(C|X_0) < c_{cut} - \epsilon \; \rightarrow \\
&EC_{A_O}(X_0) = P(C|X_0)R_{FN} \\
&EC_{A_\epsilon}(X_0) \leq (1 - P_\epsilon)P(C|X_0)R_{FN} + P_\epsilon(1-P(C|X_0))R_{FP}\\
&EAC_{A_\epsilon}(X_0) \leq P_\epsilon R_{FP} \\
&P(C|X_0) > c_{cut} + \epsilon \; \rightarrow \\
&EC_{A_O}(X_0) = (1-P(C|X_0))R_{FP} \\
&EC_{A_\epsilon}(X_0) \leq (1-P_\epsilon) (1-P(C|X_0))R_{FP} + P_\epsilon P(C|X_0)R_{FN} \\
&EAC_{A_\epsilon}(X_0) \leq P_\epsilon R_{FN} \\
&P(C|X_0) \geq c_{cut} - \epsilon \;,\; P(C|X_0) \leq c_{cut} + \epsilon \; \rightarrow \\
&EC_{A_O}(X_0) = \min \left( P(C|X_0)R_{FN} , (1-P(C|X_0))R_{FP} \right) \\
&EC_{A_\epsilon}(X_0) \leq \max \left( P(C|X_0)R_{FN} , (1-P(C|X_0))R_{FP} \right) \\
&EAC_{A_\epsilon}(X_0) \leq \epsilon (R_{FP} + R_{FN})
\end{align*}
A combined and simplified bound is:
\begin{align}
&EAC_{A_\epsilon}(X_0) \leq \max \left( \epsilon, P_\epsilon \right) \left( R_{FN} + R_{FP} \right) \\
&EAC_{A_\epsilon} \leq \max \left( \epsilon, P_\epsilon \right) \left( R_{FN} + R_{FP} \right)
\end{align}
The deviation from the optimal alarm decreases linearly with $\epsilon$ and probability of error $P_\epsilon$, but increases linearly with each error cost $R$. Using this lemma, techniques that estimate the collision probability with a certain accuracy can be guaranteed to provide accurate alarms.

\section{Techniques}
\subsection{Monte Carlo Sampling}
A straightforward method to approximate the value $P(C_t | X_0)$ is to randomly sample many positions for each vehicle at each time. The proportion of samples that include a collision is the estimate for $P(C_t | X_0)$. The full collision probability $P(C | X_0)$ can be approximated by propagating each sample across multiple timesteps and counting the samples which contained a collision at any time, as shown in algorithm 1. Monte Carlo (MC) sampling is a popular technique for vehicular collision detection \cite{mc1, mc3, other4} because of its flexibility and guaranteed accuracy if a high number of particles are used. The algorithm shown (1) is that of \cite{mc4}, albeit with multiple time steps checked as in \cite{bayesthesis}. Motion planning for vehicles is also often determined by generating a large number of random samples or potential trajectories, so MC collision detection is easily compatible \cite{mc6}. \par
\begin{figure}[t]
\label{algo1}
\begin{centering}
Algorithm 1: \hspace{2mm} MC Sampling Alarm \\
\end{centering}
current state info for all vehicles: $X_0$ \\
time resolution $\Delta_t$, cutoff time $T$, number of particles $N$ \\
$C \gets 0$ \\
\textbf{for} $i = 1, i \leq N, i$+=$1$ \\
\hphantom{1cm} $x_0 \gets$ sample from $X_0$ \\
\hphantom{1cm} \textbf{for} $t = \Delta_t , t \leq T , t$+=$\Delta_t$ \\
\hphantom{1cm} \hspace{1cm} $x_t \gets$ sample from $P\left( X_t \vert X_{t-\Delta_t} = x_{t-\Delta_t} \right)$ \\
\hphantom{1cm} \hspace{1cm} \textbf{if} Collision($x_t$) \\
\hphantom{1cm} \hspace{2cm} C += 1 \\
\hphantom{1cm} \hspace{2cm} \textbf{break} \\
\textbf{return} True \textbf{if} $C > N c_{cut}$ \\
\end{figure}
\begin{figure}[t]
\centering
\includegraphics[width=2.2in]{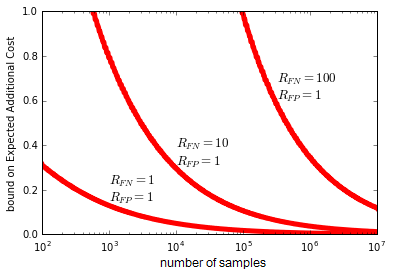}
\caption{Maximum error of an MC sampling alarm.}
\label{mcs_bound}
\end{figure}

Estimating a number using the average of independent, bounded random variables is a classic problem in probability and has been given various guarantees. The Azuma-Hoeffding lemma is one such guarantee and provides a confidence interval on the approximation error \cite{azuma}. With $\epsilon$ and $P_\epsilon$ as defined in Section II, the lemma states
\begin{equation}
P_\epsilon \leq 2 e^{-\frac{1}{2}n\epsilon^2}
\end{equation}
This bound can relate the MC sampling alarm to the approximate alarm $A_\epsilon$, which in turn allows direct calculation of the worst-case error for MC sampling. A theoretical upper limit on the Expected Additional Cost of an MC sampling alarm is shown in Figure \ref{mcs_bound} as a function of the number of samples used, with several configurations of the error penalty. Note that as the assigned cost of a missed collision increases, the required number of samples to maintain a low cost increases accordingly.

\subsection{Point-based Alarms}
The Monte Carlo method generates a large number of random possible future scenarios from the initial random set of states, $X_0$. By picking a specific set of scenarios to analyze, it may be possible to approximate the probability of collision with fewer computations. The simplest approach of this type is to take the expected position of each vehicle at each point in time and send an alarm if the expected positions include a collision \cite{mean1,mean2}.
 The expected future state of a vehicle can only be directly calculated when it follows a linear motion model. The unscented transform picks a small set of points based on the covariance of the initial probability distribution \cite{utguide}. If these points are each passed through a nonlinear function (such as the generation of future vehicle states and the check for collisions in those states), a weighted sum of their output approximates the expected value of the output. \cite{grid2} applies the unscented transform to collision detection by rewriting the probability of collision as:
\begin{equation}
P(C|X_0) = E \left[ \left\{ \begin{array}{lr} 1 & C=\textnormal{True} \\ 0 & C=\textnormal{False} \end{array} \right\} | X_0 \right]
\end{equation}
This approach requires much less computation than the MC sampling method. However, the theoretical guarantees of the unscented transform do not apply to collision detection because those guarantees rely on differentiability. The transformation from a continuous set of variables to a binary value is non-differentiable. An alternative way to use the unscented transform is to define a continuous random variable from which the occurrence of a collision is easily defined. \cite{gridut} use an unscented transform to calculate the minimum distance between cars over a period of time, reasoning that collisions occur if and only if this distance is less than zero. \cite{grid3} similarly define multiple distance measures that each represent a collision if negative. However, the unscented transform only approximates the mean and variance of a variable, not its entire distribution. In order to calculate the probability that the minimum distance between cars is less than zero, the distance must be arbitrarily assigned a distribution type, for instance truncated normal \cite{gridut}.
\par
Other alarms based on checking multiple points have been suggested. For instance, if each variable in the vehicle's state is limited to finite boundaries, the state space can be discretized into a finite number of blocks in a grid. The probability of transitioning from one block to another can be calculated in advance as in \cite{grid1}, as can the subset of blocks that represent a collision. This method scales poorly as the number of blocks in this grid will increase exponentially as the dimensionality of the state space increases. In short, a variety of alarm techniques share the framework of selecting potential vehicle positions then checking each for collision. 

\subsection{Machine Learning}
The design of optimal alarms is very similar to the machine learning task of classification, which predicts a true or false value given previous examples. These tools are generally used to determine unknown models or patterns in data, whereas collision detection can utilize already-developed predictive models. However, the techniques mentioned so far require substantial repetitive computation to accurately approximate the probability of collision at each timestep. Regression techniques can be used to generate a compact model of $P(C | X_0)$ for rapid querying, as shown in Algorithm 3. Similarly, classification can be used to directly generate alarms based on simulated collision or non-collision examples. These approaches are extremely similar, but regression is covered here because it allows for easy modification of the proper cutoff probability.\par
\begin{figure}[ht]
\label{algo3}
\begin{centering}
Algorithm 3: \hspace{2mm} Machine Learning Alarm \\
\end{centering}
\textbf{Train Model}\hrulefill \\
$D \gets N\times d$ matrix of explanatory variables \\
$c \gets N\times 1$ vector of truth \\
\textbf{for} $i=1,i\leq N,i$+=$1$ \\
\hphantom{1cm} randomly generate vehicle states $X_0$ \\
\hphantom{1cm} use a high-sample Monte Carlo alarm to find $P(C|X_0)$ \\
\hphantom{1cm} $D_i, = X_0$ , $c_i = P(C|X_0)$ \\
train model to fit $D$ to $c$ \\
\textbf{Use Model}\hrulefill \\
current state info $X_0$, optimal cutoff $c_{cut}$ \\
\hphantom{1cm} $\hat{c} =$ model prediction for $X_0$ \\
\hphantom{1cm} \textbf{return} True \textbf{if} $\hat{c} > c_{cut}$
\end{figure}

The accuracy of a regression model is often characterized by its root mean squared error, in this case $RMSE = \sqrt{E\left[ \left( \hat{c} - P(C|X_0) \right)^2 \right]}$. This value can easily be related to the expected additional cost of an approximate alarm.
\begin{align}
&\hat{c} - P(C|X_0) = \epsilon \rightarrow \vert \hat{c} - P(C|X_0) \vert = \vert \epsilon \vert \nonumber \\
&EAC_{A}(X_0) \leq \vert \epsilon \vert (R_{FN} + R_{FP}) \nonumber \\
&EAC_{A} \leq (R_{FN} + R_{FP}) E\left[ \vert \epsilon \vert \right] \nonumber \\
&E\left[ \vert \epsilon \vert \right] \leq \sqrt{E\left[ \epsilon^2 \right]} = RMSE \nonumber \\
&EAC_{A} \leq (R_{FN} +  R_{FP}) RMSE
\end{align}
There is also a wide array of modeling techniques from which to choose. The technique:
\begin{itemize}
\item Needs high flexibility or expressive power. The true function $P(C | X_0)$ is most likely complex and nonlinear.
\item Needs to rapidly predict new values. This is the primary advantage of using such models.
\item Does not need to generalize well off a small amount of data. Simulation can be used to create as many training points as necessary.
\end{itemize}
Shallow unstructured networks such as multi-layer perceptrons closely fit the description above. One disadvantage of learning techniques is that they cannot be reliably modified after training. If the model used to determine each vehicle’s current information or predict each vehicle’s position is altered in any way, or even if the cutoff time for collision detection is altered, the collision detection model must be retrained with new simulations. Another potential disadvantage is that the number of variables that describe the state $X_0$ may be quite high. For instance, not only the expected value of each variable in the vehicle's state, but their variances and correlations can be treated as explanatory variables.
\par
Machine learning algorithms have been used for collision detection in several ways. \cite{class1} used a neural network classifier to send rear-end collision warnings. \cite{class4} similarly used a Bayesian network to predict collisions in an intersection. \cite{other1} used Monte Carlo sampling to predict the probability of collision, but runs each sample through a pre-trained classifier rather than performing the complete collision check. \cite{class3} used real data to model critical events that likely preclude collisions, such as hard braking. One common detail across these works is that the model inputs are intermediate variables such as surrogates of time-to-collision, rather than raw vehicles' state information. No reasons were given for this choice, except that the calculated values are likely to have a close relationship with the probability of collision.

\begin{figure*}[h]
\centering
\subfloat[Unprotected Left Turn]{\includegraphics[width=2.9in]{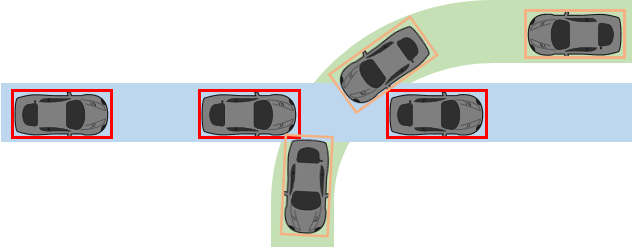}%
\label{left_turn}}
\hfil
\subfloat[Free Space with Bicycle Model]{\includegraphics[width=2.1in]{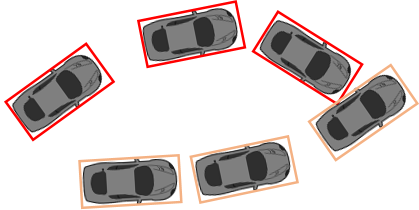}%
\label{bicycle}}
\caption{Example vehicle positions at three timesteps from the two simulated scenarios.}
\label{fig_sim}
\end{figure*}

\begin{table*}[t]
\renewcommand{\arraystretch}{1.3}
\caption{Performance of the True Optimal Alarm for each Simulation}
\label{simresults}
\centering
\begin{tabular}{l|c|c|c|c|c|c|c|c|c|c|}
\cline{2-11}
\multirow{2}{*}{} & Collision Rate & \multicolumn{3}{c|}{$R_{FN} = 1$} & \multicolumn{3}{c|}{$R_{FN} = 10$} & \multicolumn{3}{c|}{$R_{FN} = 100$} \\ \cline{3-11}  & (\%) & FNR & FPR & EC & FNR & FPR & EC & FNR & FPR & EC \\ \hline
\multicolumn{1}{|l|}{Left Turn with 1 second detection} & 4.0 & .561 & .028 & .006 & .135 & .070 & .120 & .018 & .192 & .255 \\ \hline
\multicolumn{1}{|l|}{Left Turn with 2.5 second detection} & 7.1 & .870 & .006 & .068 & .221 & .174 & .319 & .018 & .561 & .646 \\ \hline
\multicolumn{1}{|l|}{Bicycle Model with 1 second detection} & 34 & .252 & .102 & .154 & .021 & .441 & .361 & .001 & .663 & .480 \\ \hline
\end{tabular}
\end{table*}

\section{Experiments}
The performance and speed of several alarm techniques were tested in two simulated scenarios, each involving two vehicles. Each vehicle is shaped as a 5m by 2m rectangle and moves according to a discrete-time Markov process with Gaussian error. The time resolution is ten steps per second. The vehicles are positioned randomly within ten meters of each other, and are then moved backwards in time to reach an initial position for the simulation. This ensures that the simulations contain a variety of collision and near-collision situations. Collision detection is only applied at the starting time of the simulation to provide distinct cases of correct or incorrect alarms. In reality, collision detection would take place at regular intervals, and alarms could be judged as not only correct or incorrect but also late or early
\par
The first scenario is an unprotected left turn conflict as in \cite{mc5}, shown in Figure \ref{left_turn}. The vehicles are each fixed to a path through the intersection, meaning they can only move forward or backward. Their motion follows a nearly-constant velocity model as in \cite{grid1}. Two separate sets of one thousand simulations each were run, the first detecting collisions up to one second in advance and second detecting collisions up to 2.5 seconds. In the second simulated scenario, both vehicles move in unbounded 2D space according to the bicycle model, described in \cite{bike1,mcplan} among others. Each vehicle's state is given by six parameters: $x$ and $y$ position, orientation angle, velocity, acceleration, and angular velocity. One thousand simulations were run for collision detection of up to one second. \par
Table \ref{simresults} shows the performance of an optimal alarm on each simulated scenario. The optimal alarm was achieved by a Monte Carlo alarm with 20000 samples, which is too slow for practical use but highly accurate. The probability cutoff at which to send an alarm was chosen as described in Section II, with the false positive cost set to 1 and the false negative cost set to 1, 10, and 100. For each of these individual costs, the alarm's performance is given in terms of false negative rate (FNR), false positive rate (FPR), and expected cost (EC).\par

\begin{table}[t]
\renewcommand{\arraystretch}{1.3}
\caption{Approximate Alarm Performance}
\label{simresults1}
\begin{tabular}{|l|c|c|c|c|}
\hline 
\hphantom{5mm} Left Turn & Runtime & \multicolumn{3}{c|}{Expected Additional Cost} \\
\cline{3-5} 1 second detection & (avg) (ms) & $R_{FN}=1$ & $10$ & $100$ \\ \hline
MC 10 samples & 6 & .002 & .031 & .387 \\ \hline
MC 100 samples & 8 & .000 & .002 & .029 \\ \hline
MC 1000 samples & 23 & .000 & .000 & .004 \\ \hline
9 unscented samples & 5 & .001 & .009 & .010 \\ \hline 
Expected Value & 2 & .001 & .089 & 1.76 \\ \hline
MLP Regression & 0.1 & .027 & .034 & .195 \\ \hline
\hphantom{5mm} Left Turn &  & \multicolumn{3}{c|}{} \\
2.5 second detection &  & \multicolumn{3}{c|}{} \\ \hline
MC 10 samples & 16 & .003 & .066 & .930 \\ \hline
MC 100 samples & 19 & .000 & .006 & .070 \\ \hline
MC 1000 samples & 43 & .000 & .001 & .010 \\ \hline
9 unscented samples & 12 & .002 & .017 & .027 \\ \hline
Expected Value & 6 & .002 & .260 & 5.03 \\ \hline
MLP Regression & 0.1 & .002 & .182 & .201 \\ \hline
\hphantom{1mm} Bicycle Model &  & \multicolumn{3}{c|}{} \\ 
1 second detection &  & \multicolumn{3}{c|}{} \\ \hline
\multicolumn{1}{|l|}{MC 10 samples} & 6 & .009 & .035 & .562 \\ \hline
\multicolumn{1}{|l|}{MC 100 samples} & 8 & .002 & .006 & .026 \\ \hline
\multicolumn{1}{|l|}{MC 1000 samples} & 23 & .000 & .001 & .002 \\ \hline
\multicolumn{1}{|l|}{73 unscented samples} & 5 & .130 & 1.04 & 11.4 \\ \hline
\multicolumn{1}{|l|}{Expected Value} & 2 & .021 & .659 & 8.99 \\ \hline
\multicolumn{1}{|l|}{MLP Regression} & 0.1 & .190 & .219 & .195 \\ \hline
\end{tabular}
\end{table}

The approximate alarm techniques tested are: Monte Carlo alarms with 100, 1000, and 10000 samples, a point-based alarm using only the expected value of each vehicle's state, a point-based alarm using a three-layer unscented transform (for each vehicle), and a machine learning alarm using a 150-node multilayer perceptron regressor, trained with one million simulations. The performance of each alarm for each simulated scenario is shown in Table \ref{simresults1} in terms of expected additional cost. Calculations were performed with Python 2.7 on a single computer. Operations were vectorized using Numpy for speed. The machine learning alarm utilizes the neural\_network package from Sci-kit Learn and thus is already optimized for speed. A lower-level programming language could lower the runtime of some techniques but is not expected to change the ordering of algorithms by runtime as shown here. The simulation code is available at https://github.com/utexas-ghosh-group/carstop/VCD. \par

MC sampling alarms with 100 and 1000 samples achieve accuracy fairly close to optimal alarms, despite their theoretical maximum error being quite high as shown in Section III. The runtime of low-sample MC alarms was on the order of ten milliseconds, whereas multilayer perceptron regression can send alarms in less than a millisecond but with much poorer accuracy. The unscented transform worked well on the left turn scenario but very poorly in the more complex and nonlinear bicycle scenario.

\section{Conclusion}
This paper provides a common framework for describing and comparing several methods of probabilistic vehicular collision detection. Monte Carlo sampling is known to asymptotically achieve the correct response, but its relative real-time performance had not previously been quantified. Probabilistic bounds and experiments were used to quantify the number of samples needed for a certain level of robustness. The experiments showed that Monte Carlo sampling is a highly competitive technique except when millisecond-order speed is desired. Which faster approximation is appropriate may depend on the complexity of the model. 

\section*{Acknowledgment}
This work was supported by the Texas Department of Transportation under Project 0-6877 entitled ``Communications and Radar-Supported Transportation Operations and Planning (CAR-STOP).''


\bibliographystyle{IEEEtran}
\bibliography{IEEEabrv,main}
\end{document}